%% file: main.tex
\documentclass[10pt,twocolumn,letterpaper]{article}

\usepackage{adjustbox}

\usepackage{iccv}              


\definecolor{iccvblue}{rgb}{0.21,0.49,0.74}
\usepackage[pagebackref,breaklinks,colorlinks,allcolors=iccvblue]{hyperref}

\def\paperID{*****} 
\def\confName{ICCV}
\def\confYear{2025}

\title{Tool-Augmented Agent for Closed-loop Optimization, Simulation, and Modeling Orchestration}

\author{
Liyuan Deng\textsuperscript{1,2}\thanks{L.~Deng and S.~Deng contributed equally.} \quad 
Shujian Deng\textsuperscript{2}\footnotemark[1] \quad 
Yongkang Chen\textsuperscript{2} \quad 
Yongkang Dai\textsuperscript{1} \quad 
Zhihang Zhong\textsuperscript{2} \quad \\ 
Linyang Li\textsuperscript{2} \quad 
Xiao Sun\textsuperscript{2} \quad 
Yilei Shi\textsuperscript{1}\thanks{Y.~Shi and H.~Huang are co-corresponding authors.} \quad 
Huaxi Huang\textsuperscript{2}\footnotemark[2] \\
\textsuperscript{1}Northwestern Polytechnical University \quad 
\textsuperscript{2}Shanghai Artificial Intelligence Laboratory \\
{\tt\small \{dly,yilei\_shi\}@mail.nwpu.edu.cn} \quad
{\tt\small \{dengshujian,huanghuaxi\}@pjlab.org.cn}
}

\begin{document}
\twocolumn[{
    \begin{center}
        \renewcommand\twocolumn[1][]{#1}
        
        \vspace*{-2.5cm} 
        {\maketitle}
  
        \vspace{-1.5em} 

        \centering
        \includegraphics[width=\textwidth]{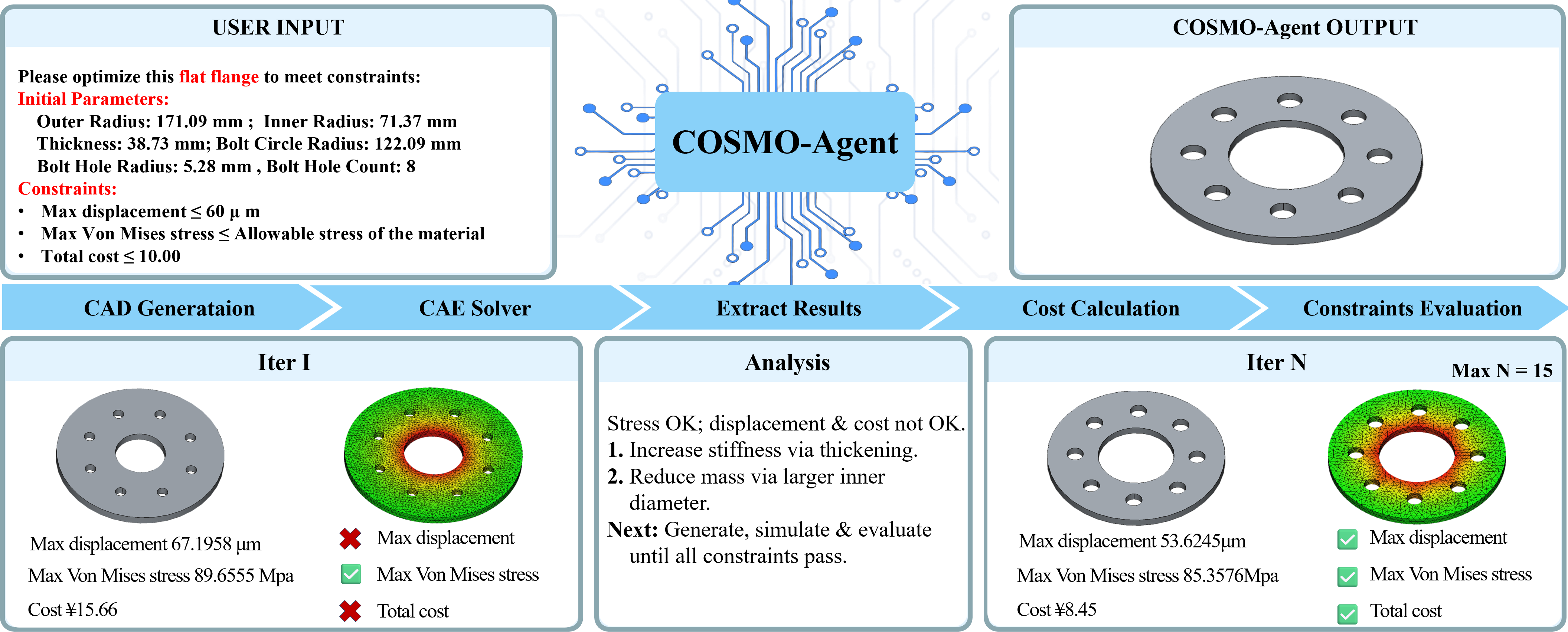}
        
        \vspace{0.3em} 
        
        \begin{minipage}{\textwidth} 
            \centering
            \captionsetup{type=figure} 
            \caption{COSMO-Agent performs closed-loop CAD--CAE optimization by iteratively generating parametric geometry, running CAE simulation, extracting displacement/stress metrics, and updating design parameters until all constraints are satisfied.}
            \label{fig:teaser}
        \end{minipage}
        
        \vspace{-1em} 
    \end{center}
}]
\footnotetext[1]{L.~Deng and S.~Deng contributed equally.}
\footnotetext[2]{Y.~Shi and H.~Huang are co-corresponding authors.}
      
\input{sec/0_abstract}    
\input{sec/1_intro}
\input{sec/2_formatting}

\input{sec/3_method}

\input{sec/exp}
\input{sec/conclusion}

{
    \small
    \bibliographystyle{ieeenat_fullname}
    \bibliography{main}
}


\end{document}

%% file: sec/0_abstract.tex
\begin{abstract}
\vskip -1.0em  
Iterative industrial design–simulation optimization is bottlenecked by the CAD–CAE semantic gap: translating simulation feedback into valid geometric edits under diverse, coupled constraints. To fill this gap, we propose \textbf{COSMO-Agent} (\textbf{C}losed-loop \textbf{O}ptimization, \textbf{S}imulation, and \textbf{M}odeling \textbf{O}rchestration), a tool-augmented reinforcement learning (RL) framework that teaches LLMs to complete the closed-loop CAD–CAE process. Specifically, we cast CAD generation, CAE solving, result parsing, and geometry revision as an interactive RL environment, where an LLM learns to orchestrate external tools and revise parametric geometries until constraints are satisfied. To make this learning stable and industrially usable, we design a multi-constraint reward that jointly encourages feasibility, toolchain robustness, and structured output validity. In addition, we contribute an industry-aligned data set that covers 25 component categories with executable CAD–CAE tasks to support realistic training and evaluation. Experiments show that COSMO-Agent training substantially improves small open-source LLMs for constraint-driven design, exceeding large open-source and strong closed-source models in feasibility, efficiency, and stability.
\end{abstract}

%% file: sec/1_intro.tex
\section{Introduction} \label{intro}
Modern industrial design is an iterative search for geometries that satisfy coupled and often competing constraints. In functional component development, Computer-Aided Design (CAD) specifies a parametric geometry with a feature/history tree, while Computer-Aided Engineering (CAE) provides physics-based verification through simulation (e.g., finite element analysis). Despite advances in automation, \emph{closed-loop} CAD--CAE iteration remains a practical bottleneck: engineers must translate high-dimensional simulation feedback (fields and aggregate metrics) into low-dimensional, \emph{structured} CAD edits that remain executable under the original parametric history. This translation is further complicated by heterogeneous toolchains that frequently fail mid-pipeline, where regeneration breakdowns, meshing errors, and solver non-convergence are common.
As a result, CAD--CAE optimization in practice becomes a long-horizon sequential decision-making problem under hard executability constraints and stochastic tool failures, rather than a clean continuous optimization problem.

Existing automation strategies only partially address this setting. Derivative-free optimizers \cite{jones1998ego,hansen2001cmaes,audet2006mads,snoek2012practicalbo} can adjust parameters to scalar objectives, but typically do not model \emph{executability} and \emph{failure recovery} as part of the optimization state. Moreover, validity is usually enforced by rigid templates or hand-crafted heuristics.
Differentiable or surrogate-based methods can reduce expensive solver calls, but frequently rely on approximations that diverge from production CAD--CAE pipelines and do not directly produce history-consistent executable edits in native parametric CAD \cite{raissi2019pinns,lu2021deeponet,li2021fno,hu2020difftaichi}. Recent progress suggests an alternative: an LLM that serves as a controller, mapping tool feedback to structured tool invocations \cite{yao2023react,ahn2022saycan}, while tool-use training improves API-call fidelity \cite{schick2023toolformer,qin2024toollm}.
However, prompting-first agents remain brittle under regeneration/meshing/solver failures, while standard instruction tuning or RLHF mainly targets short-horizon imitation/alignment rather than long-horizon trial-and-error optimization driven by downstream simulation consequences \cite{wei2022chain,wang2023selfinstruct,ouyang2022instructgpt,christiano2017preferences}.

To address these challenges, we introduce \textbf{COSMO-Agent} (\textbf{C}losed-loop \textbf{O}ptimization, \textbf{S}imulation, and \textbf{M}odeling \textbf{O}rchestration), a novel tool-augmented reinforcement learning framework for reliable closed-loop CAD--CAE iterative evolution. We model CAD editing, regeneration, meshing, solving and result parsing as an interactive environment with explicit failure states. An LLM policy operates over a structured action space of parametric edits. Each proposed edit is validated by a CAD tool set and evaluated through simulation, and the agent iteratively revises actions based on tool feedback until all constraints are satisfied or the budget is exhausted. To make learning stable in partially reliable pipelines, we optimize a multi-constraint objective that jointly encourages (i) \emph{feasibility} via constraint satisfaction, (ii) \emph{robustness} via successful execution and recovery from toolchain failures, and (iii) \emph{structured validity} aligned with parametric requirements, preventing reward hacking that produces numerically favorable but non-executable designs.

Finally, we contribute an industry-aligned benchmark of $\sim$20,000 executable CAD--CAE tasks across 25 component categories, with standardized interfaces and fixed tool-call/retry budgets. Each task provides an initial parametric CAD model, a toolchain configuration, and constraints spanning physics, geometry, and economics (e.g., cost), enabling reproducible evaluation of feasibility, efficiency (iterations/tool calls), and stability (failure recovery). Using this benchmark, we compare diverse open-source and proprietary LLMs under a unified interface and fixed budgets. Results show that COSMO-Agent training markedly improves an 8B LLM, achieving higher feasibility, efficiency, and stability than most baselines under our protocol.
\textbf{In summary, our contributions are as follows:}
\begin{itemize}
\item formulating closed-loop CAD--CAE iterative evolution as a long-horizon sequential decision-making problem that explicitly models heterogeneous tools, hard executability constraints, and stochastic failure states.
\item proposing COSMO-Agent, a tool-augmented RL framework with a multi-constraint objective that grounds structured, executable parametric edits in downstream feedback, jointly optimizing feasibility, robustness to tool failures, and output validity.
\item introducing an industry-aligned executable CAD--CAE benchmark with standardized interfaces, fixed tool-call and retry budgets, and constraints (i.e., physics, geometry, and cost). 
\item demonstration of improved closed-loop performance under these controlled budgets.
\end{itemize}

%% file: sec/2_formatting.tex
\section{Related Work} \label{r_wks}

\subsection{CAD Model Generation}
Learning-based research on \emph{parametric} CAD spans representations and generation strategies. SketchGraphs \cite{seff2020sketchgraphs} provides large-scale constraint graphs from real CAD sketches. Fusion 360 Gallery \cite{willis2021fusion360gallery} introduces a programmatic CAD language with human design sequences and an interactive environment that formulates CAD construction as a sequential decision process. JoinABLe \cite{willis2022joinable} extends learning to CAD assemblies by releasing weakly supervised joint annotations.

Recent work also leverages large models to synthesize or manipulate CAD \emph{programs}. LLM4CAD \cite{li2024llm4cad} studies multimodal LLMs for generating CAD programs from text and images, while Text-to-CadQuery \cite{xie2025texttocadquery} directly generates CadQuery code and improves executability and geometric quality via supervision and fine-tuning. OpenECAD \cite{yuan2024openecad} enables editable CAD through structured sketches and executable construction commands, and tool-augmented agents such as CAD-Assistant \cite{mallis2025cadassistant} iteratively execute and repair CAD commands via a CAD API. Overall, prior systems largely emphasize geometric correctness, editability, or task completion, but rarely incorporate downstream CAE feedback and engineering acceptance constraints into a closed-loop objective, or treat executability and failure recovery in real CAD pipelines as first-class optimization targets.


\begin{figure*}[t]
  \centering
  \includegraphics[width=\textwidth]{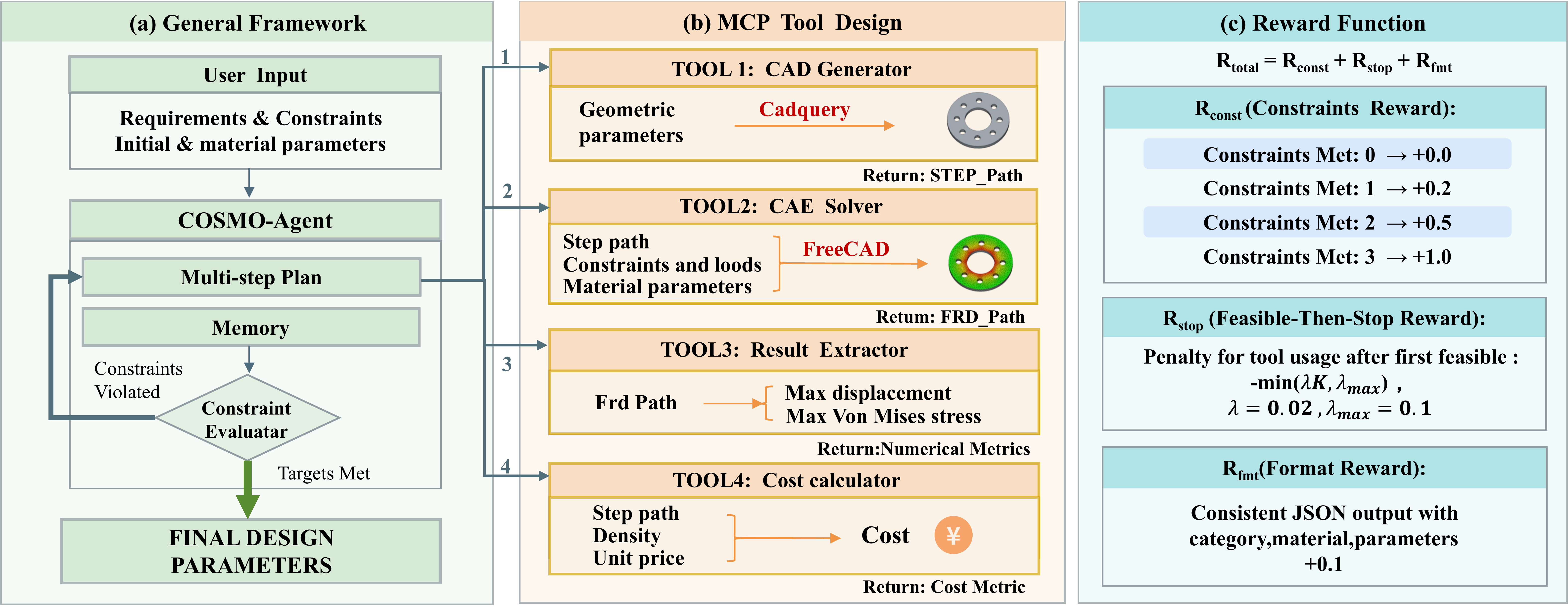}
  \caption{COSMO-Agent: (a) overall closed-loop framework, (b) MCP tool set for CAD–CAE optimization, and (c) training reward function.}
  \label{fig:framework}
\end{figure*}

\subsection{LLM Agents for Engineering Simulation}
A growing body of work couples LLMs with engineering simulators to automate solver setup and execution. In CFD, MetaOpenFOAM \cite{chen2024metaopenfoam} adopts a multi-agent architecture for OpenFOAM workflows, often using retrieval for configuration generation and error correction; CFDagent \cite{xu2025cfdagent} similarly decomposes preprocessing, solving, and postprocessing into specialized agents with iterative debugging. Other efforts build training data and fine-tuned models for natural language to solver configuration, such as NL2FOAM \cite{dong2025nl2foam}; recent end-to-end automation continues this direction (e.g., Foam-Agent \cite{yue2025foamagent}), and finite-element workflows have also been explored in ecosystems such as MOOSE \cite{zhang2025mooseagent}.
These systems demonstrate the promise of ``LLM + tools'' for generating simulation inputs, debugging configurations, and producing results. However, they largely target completing (or reproducing) a simulation instance from a specification. For multi-round design optimization—iteratively editing geometry based on simulation outcomes until multiple coupled acceptance constraints are met—learnable closed-loop strategies remain underexplored, especially under realistic toolchain instability.


\subsection{Tool-augmented LLM Agents}
For LLM agents, tool augmentation grounds decisions in executable actions and enables policy updates from observable tool feedback, which is crucial for multi-step tasks with external dependencies. ReAct \cite{yao2023react}, MRKL \cite{karpas2022mrkl}, and SayCan \cite{ahn2022saycan} exemplify reasoning interleaved with tool calls, while tool-use training improves call timing and API-call fidelity \cite{schick2023toolformer,qin2024toollm}.

Beyond prompting, recent work scales long-horizon training and optimization via verifiable feedback and efficient rollouts. InternBootcamp \cite{li2025internbootcamptechnicalreportboosting} provides verifiable task environments to support scalable RL and evaluation, HybridFlow \cite{sheng2024hybridflow} improves RLHF system efficiency for multi-step behaviors, and MARTI \cite{marti2025} unifies multi-agent training and inference with multi-turn rollouts and verifier-based workflows. However, these frameworks do not directly resolve closed-loop CAD--CAE optimization, where agents must produce structured, history-consistent parametric edits under hard executability constraints and fixed tool-call/retry budgets while remaining robust to stochastic toolchain failures. COSMO-Agent addresses this gap by training an LLM policy with explicit failure states and a multi-constraint objective grounded in downstream CAE feedback and engineering acceptance constraints.

%% file: sec/3_method.tex
\section{Methodology} \label{method}
As shown in Fig.~\ref{fig:framework}, we introduce the COSMO-Agent from three aspects: 1) the general framework, 2) the MCP tool design, and 3) the reward module. 

\subsection{General Framework}
The general framework of COSMO-Agent is as shown in Fig.~\ref{fig:framework}(a). We construct design requirements, constraints, initial geometric parameters and material parameters given by users into a prompt and input it into the LLM. We denote each task instance as: 
\begin{equation}
\mathcal{I}=\big(c,\mathbf{p}_0,\eta,\delta,\gamma,\kappa,\mathcal{M}\big)
\end{equation}
where $c$ denotes the part category, $\mathbf{p}_0\in\mathbb{R}^d$ is the initial geometric parameter vector, $\eta$ specifies the simulation settings (e.g., loads and boundary conditions), $\delta$ is the maximum displacement threshold, $\gamma$ is the maximum allowable von Mises stress threshold, $\kappa$ is the cost threshold, and $\mathcal{M}$ is the material library, with the stress limit being material-dependent and provided by the library, i.e., $\gamma(m_t)=\sigma_{\text{allow}}(m_t)$.
The LLM then formulates a round-by-round updated design plan $\{(\mathbf{p}_t,m_t)\}_{t=0}^{T}$ based on the input prompt.
At each round $t$, the design state is determined by geometric parameters $\mathbf{p}_t$ and a material choice $m_t\in\mathcal{M}$. We then feed the design state into the LLM to get the invocation strategy for MCP tool as follow:
\begin{equation}
x_t=(c,\mathbf{p}_t,m_t)
\end{equation}
The planner then invokes the MCP Tools. These MCP Tools generate three-dimensional CAD design files, and the CAE solver performs simulation solving and calculation in accordance with the specified conditions. The MCP tool returns, under simulation setting $\eta$, a scalar feedback tuple:
\begin{equation}
\Phi(x_t;\eta)=\big(u_{\max}^{(t)},\sigma_{\max}^{(t)},C^{(t)}\big)
\end{equation}
where $u_{\max}^{(t)}$ is the maximum displacement magnitude, $\sigma_{\max}^{(t)}$ is the maximum von Mises equivalent stress, and $C^{(t)}$ is the cost metric. The results are then fed back to the LLM for verifying whether they meet the design requirements. Design feasibility is defined by the following constraints:
\begin{equation}
u_{\max}^{(t)}\le \delta,\qquad
\sigma_{\max}^{(t)}\le \sigma_{\text{allow}}(m_t),\qquad
C^{(t)}\le \kappa
\end{equation}
where $\sigma_{\text{allow}}(m_t)$ denotes the allowable stress of material $m_t$ from the material library. If the requirements are not met, the planner re-initiates another CAD–CAE iteration. At each round, it records the user input and interaction history in memory, evaluates constraints using the latest $\Phi(x_t;\eta)$, and outputs updated design parameters and material choice $(\mathbf{p}_{t+1}, m_{t+1})$. The update is guided by numerical feedback: displacement and stress are parsed from CAE results, and cost is computed from geometry and material properties. The model then decides the next modification toward satisfying all constraints and triggers the next CAD generation and CAE verification. Once all constraints are satisfied at some round $t^{*}$, it returns the final design parameters/material configuration and terminates; otherwise, it proceeds until reaching the maximum number of rounds.


\subsection{MCP Tool Set}

As shown in Fig.~\ref{fig:framework}(b), we expose the CAD--CAE toolchain via MCP, allowing COSMO-Agent to trigger external computations and receive outputs through a unified, structured interface. The tool set covers four stages of the closed-loop iteration: (i) parametric geometry generation, (ii) finite element solving, (iii) result metric extraction, and (iv) cost estimation.

\paragraph{Tool 1: CAD generator.}
The CAD generator maps a part category and a parameter vector to an executable solid geometry consumable by downstream solvers. Given category $c$ and parameters $\mathbf{p}$, the tool produces a solid $G(c,\mathbf{p})$ and exports a geometry file path as the primary output. It also returns geometry metadata for boundary-condition assignment, such as anchor points on designated functional faces, so that the semantics of loads and constraints remain consistent across different parameter settings.


\paragraph{Tool 2: CAE solver.}
The CAE solver performs physics-based analysis on the generated geometry. The tool takes as input the geometry file path, material parameters (e.g., Young's modulus), and load/boundary-condition parameters (e.g., pressure magnitude and fixed constraints), and outputs a standard result file path together with solver logs. To enable automated boundary-condition assignment, anchor points in the geometry metadata are matched to target faces after geometry import. Let the candidate face set be
\begin{equation}
\mathcal{F}=\{F_j\}
\end{equation}
and for an anchor point $\mathbf{q}$, the point-to-face distance is $\operatorname{dist}(\mathbf{q},F_j)$. Faces satisfying
\begin{equation}
\operatorname{dist}(\mathbf{q},F_j)\le \varepsilon
\end{equation}
within tolerance $\varepsilon$ are selected as load faces or constraint faces, which maintains consistent boundary-condition locations across varying parameterizations.


\paragraph{Tool 3: Result extractor.}
The result extractor converts field outputs in the solver result file into scalar metrics used for constraint checking. Its input is the result file path produced by the CAE solver, and its outputs are the maximum displacement $u_{\max}$ and the maximum equivalent stress $\sigma_{\max}$. For the nodal displacement vector $\mathbf{u}_i=(u_{x,i},u_{y,i},u_{z,i})$, the displacement magnitude and maximum displacement are defined as
\begin{equation}
\|\mathbf{u}_i\|_2=\sqrt{u_{x,i}^2+u_{y,i}^2+u_{z,i}^2},\\
u_{\max}=\max_i\|\mathbf{u}_i\|_2
\end{equation}
For the stress components $(\sigma_{xx},\sigma_{yy},\sigma_{zz},\tau_{xy},\tau_{yz},\tau_{zx})$, we define
\begin{equation}
\begin{aligned}
\Delta_\sigma &= (\sigma_{xx}-\sigma_{yy})^2 + (\sigma_{yy}-\sigma_{zz})^2 + (\sigma_{zz}-\sigma_{xx})^2, \\
\Delta_\tau   &= \tau_{xy}^2 + \tau_{yz}^2 + \tau_{zx}^2
\end{aligned}
\end{equation}
The von Mises equivalent stress is then computed as
\begin{equation}
\sigma_v=\sqrt{\tfrac{1}{2}\Delta_\sigma + 3\Delta_\tau}
\end{equation}
and the maximum equivalent stress is computed by
\begin{equation}
\sigma_{\max}=\max_i \sigma_{v,i}
\end{equation}
These outputs provide numerical observations for constraint evaluation and iterative updates.


\paragraph{Tool 4: Cost calculator.}
The cost calculator provides a cost metric associated with geometry and material. Its inputs include the geometry file path, the material density $\rho(m)$, and the unit mass price $\pi(m)$, and its output is the cost $C$. The computation follows a volume--mass--price chain. Let the solid volume obtained from the geometry be $V_{\text{m}^3}$ (in $\text{m}^3$). The mass is computed as
\begin{equation}
M=\rho(m)\,V_{\text{m}^3}
\end{equation}
and the final cost is
\begin{equation}
C=M\cdot \pi(m)
\end{equation}
Along with displacement and stress, this cost is used to check the cost constraint and guide subsequent updates.

\subsection{Reward Design}
We continue training from Qwen3-8B \cite{yang2025qwen3technicalreport} so that the model can learn tool-usage strategies over multi-round interactions and produce structured outputs that are directly executable. We adopt Generalized Reinforcement Policy Optimization (GRPO)\cite{shao2024deepseekmathpushinglimitsmathematical} to optimize the policy, where the reward signal aligns multiple objectives within a single training target, including satisfying engineering constraints, reducing redundant tool invocations, and maintaining consistency between the reported outputs and the executed design state, as illustrated in Fig.~\ref{fig:framework}(c).

The reward is derived by parsing the rollout tool-interaction logs, without performing additional CAE re-evaluation. Reward computation relies solely on the tool calls, tool responses recorded in the trajectory, and the model’s final output, thereby ensuring that the training-time evaluation remains consistent with the actual execution of the toolchain. The overall reward consists of three components:
\begin{equation}
R = R_{\text{cons}} + R_{\text{stop}} + R_{\text{json}}
\end{equation}


\paragraph{Constraint Reward $R_{\text{cons}}$.}
From the trajectory log, we extract the metric triple $(u_{\max},\sigma_{\max},C)$ for each iteration, where $u_{\max}$ and $\sigma_{\max}$ are taken from the result extractor outputs and $C$ is taken from the cost calculator output. The material $m$ and geometric parameters $\mathbf{p}$ are taken from the most recent design proposal associated with that triple. We use the last complete triple in the trajectory as the final evaluation target; if no complete triple exists, we set $R_{\text{cons}}=0$. For the final triple, we define the number of satisfied constraints as
\begin{equation}
N=\mathbb{I}[u_{\max}\le \delta]+\mathbb{I}[C\le \kappa]+\mathbb{I}[\sigma_{\max}\le \sigma_{\text{allow}}(m)]
\end{equation}
The constraint reward is defined in a piecewise manner:
\begin{equation}
R_{\text{cons}}=
\begin{cases}
0.00,& N=0\\
0.20,& N=1\\
0.50,& N=2\\
1.00,& N=3
\end{cases}
\end{equation}

\paragraph{Feasible-Then-Stop Term $R_{\text{stop}}$.}
We locate the first time step at which a complete triple satisfying all three constraints appears in the trajectory, and denote its time index by $t_{\text{feas}}$. Let $K$ be the number of tool events after $t_{\text{feas}}$ (including tool calls and tool responses). We define
\begin{equation}
R_{\text{stop}}=-\min(\lambda K,\lambda_{\max})
\end{equation}
where $\lambda=0.02$ and $\lambda_{\max}=0.10$. If no complete triple satisfying all constraints appears in the trajectory, we set $R_{\text{stop}}=0$.

\paragraph{Structured-Output Consistency Term $R_{\text{fmt}}$.}
To support downstream CAD generation and simulation reproducibility, the final output is required to be a structured JSON object (including category, material, and geometric parameters). If the final output contains a parsable JSON whose category/material/parameters are consistent with the design proposal that produced the final triple, we assign
\begin{equation}
R_{\text{fmt}}=0.10
\end{equation}
otherwise, we set
\begin{equation}
R_{\text{fmt}}=0
\end{equation}


\section{Dataset Annotation}\label{sec:dataset_annotation}

To better support the CAD--CAE closed-loop optimization, we construct an industrial component dataset. The dataset covers 25 common part categories in industrial design, including flat plate flanges, triangular brackets, hex thin nuts, and I-beam cantilever beams. In terms of scale, it contains 20,000 training samples, 200 test samples, and 100 generalization samples. Among the 25 categories, 20 are used for training and testing, while the remaining 5 are reserved for generalization evaluation. The generalization set is designed to assess the model's transfer ability to unseen geometric templates and parameter semantics.

The geometric data are generated from parametric templates. For each part category, we implement a CadQuery \cite{cadquery_contributors_2025_14590990} template that produces a STEP file given a set of geometric parameters. We then perform finite element analysis under the specified material and boundary/loading conditions, and extract the maximum displacement magnitude $u_{\max}$ and the maximum von Mises equivalent stress $\sigma_{\max}$. The cost metric $C$ is computed from the geometry volume together with the material density and unit price. Material properties are provided by a unified material library $\mathcal{M}$, including $E,\nu,\rho,\pi,$ and $\sigma_{\text{allow}}$, summarized in Table~\ref{tab:material_lib}. In particular, $\sigma_{\text{allow}}$ is directly used as the upper bound in stress constraint checking.

To construct optimization-oriented targets, we annotate the constraint thresholds based on the ground-truth metrics obtained from simulation. For displacement and cost, we adopt a randomized reduction strategy: a standard reduction of 5\%--10\%, and an extreme reduction of 30\% (accounting for 10\% of the dataset). We further vary the difficulty by applying the reduction to one, two, or all three constraints (``reduce 1 / 2 / 3 items''), thereby producing a diverse set of constraint combinations. After threshold generation, we perform a feasibility check to ensure that the target thresholds remain within the feasible range for the corresponding category, avoiding trivially infeasible targets.

Beyond numerical annotations, we also construct prompts to drive interactive optimization. Each prompt is formed by concatenating four parts in order: background and objectives (Part~1), initial design and boundary/loading description (Part~2), material library, tool usage rules, and termination conditions (Part~3), and the fixed JSON output requirement (Part~4). Parts~1 and~3 each include 10 stylistically different variants to increase linguistic diversity, while Parts~2 and~4 are customized for each category to ensure strict consistency in parameter semantics, boundary conditions, and JSON field definitions. Finally, the model is required to output a single parsable JSON object containing the optimized geometric parameters and the selected material name, enabling reproducible CAD generation and CAE verification.

\begin{table}[t]
\centering
\caption{Material library used in our CAD--CAE toolchain. For each material, $E$ denotes Young's modulus in MPa, $\nu$ denotes Poisson's ratio, $\rho$ denotes mass density in kg/m$^{3}$, $\pi$ denotes unit mass price in \textyen/kg, and $\sigma_{\text{allow}}$ denotes the allowable stress in MPa.}
\label{tab:material_lib}
\begin{adjustbox}{width=\columnwidth,center}
\begin{tabular}{l *{5}{r}}
\hline
\multicolumn{1}{c}{\textbf{Name}} &
\multicolumn{1}{c}{\textbf{$E$}} &
\multicolumn{1}{c}{\textbf{$\nu$}} &
\multicolumn{1}{c}{\textbf{$\rho$}} &
\multicolumn{1}{c}{\textbf{Price}} &
\multicolumn{1}{c}{\textbf{$\sigma_{\text{allow}}$}} \\
\hline
Carbon Steel - ASTM A105           & 210000 & 0.30 & 7900 & 6.0  & 167 \\
Stainless Steel 304                & 193000 & 0.29 & 8000 & 16.0 & 137 \\
ASTM A333 Gr.6                     & 202000 & 0.30 & 7850 & 8.0  & 160 \\
Gray Cast Iron                     & 110000 & 0.25 & 7200 & 8.0  & 200 \\
Chrome-Moly Alloy Steel            & 203000 & 0.29 & 7800 & 11.0 & 300 \\
\hline
\end{tabular}
\end{adjustbox}
\end{table}

%% file: sec/exp.tex
\section{Experiments} \label{exps}
\subsection{Experiment Settings}
\subsubsection{Implementation Details}
COSMO-Agent is built on the Qwen3-8B backbone and trained with the Internbootcamp \cite{li2025internbootcamptechnicalreportboosting} framework for multi-turn interactive rollouts and policy updates on 16$\times$H200 (144GB) GPUs. We use GRPO during training: for each prompt, we sample 8 rollout trajectories with temperature=1.0 and top-$p$=0.9; each instance allows up to 15 assistant turns to cover multi-round iterations in the CAD--CAE closed loop. For optimization, we set the actor learning rate to $1\times 10^{-6}$, adopt GRPO clipping with clip ratio low=0.2 and high=0.28, and apply gradient clipping with a norm of 1.0. The training batch size is 8 prompts per step, and dynamic batching is enabled to accommodate length variability from long contexts and multi-turn interactions. We also enable KL regularization to constrain policy deviation, with the KL coefficient set to 0.001. We choose CADQuery library as the CAD generator. The CAE solver performs finite element analysis using the FreeCAD \cite{FreeCAD} FEM backend, where meshing is carried out by Gmsh \cite{geuzaine2009gmsh} and linear static solving is performed by CalculiX.

\subsubsection{Compared Methods}
We compare COSMO-Agent with language models of different scales, covering both state-of-the-art open-source and closed-source systems. All models are evaluated under the same task inputs, the same CAD--CAE toolchain, the same maximum interaction-turn limit, and a unified JSON output specification. The open-source baselines include Qwen3-8B \cite{yang2025qwen3technicalreport}, Intern-S1-mini \cite{bai2025interns1scientificmultimodalfoundation}, Llama-4-Scout \cite{Llama4Scout}, Qwen3-30B \cite{qwen3technicalreport}, Qwen3-Next \cite{qwen3technicalreport}, and Intern-S1 \cite{bai2025interns1scientificmultimodalfoundation}. The closed-source baselines include Claude-Sonnet-4.5 \cite{ClaudeSonnet45SystemCard} and Gemini-3-Flash \cite{Gemini3FlashModelCard}.

\subsubsection{Evaluation Metrics}
We evaluate all models in terms of task success and efficiency. For each instance, the model's final JSON output is parsed into executable geometric parameters and a material configuration. We then reproduce the result with the same CAD--CAE toolchain to obtain the maximum displacement magnitude $u_{\max}$, the maximum equivalent stress $\sigma_{\max}$, and the cost $C$, based on which we compute the following metrics:

\noindent$\bullet$~\textbf{Full Success Rate} (FSR): the proportion of instances that satisfy all three constraints (displacement, stress, and cost).\\
$\bullet$~\textbf{Displacement Satisfaction Rate} (DSR): the proportion of instances with $u_{\max}\le \delta$.\\
$\bullet$~\textbf{Stress Satisfaction Rate} (SSR): the proportion of instances with $\sigma_{\max}\le \sigma_{\text{allow}}(m)$.\\
$\bullet$~\textbf{Cost Satisfaction Rate} (CSR): the proportion of instances with $C\le \kappa$.\\
$\bullet$~\textbf{Model Extract Output} (MEO): the proportion of instances for which a valid and parsable final JSON can be successfully extracted from the model output.\\
$\bullet$~\textbf{Average Score} (AS): the average per-instance composite score, consisting of (i) success/failure signals from tool responses, (ii) whether a valid structured JSON can be extracted, and (iii) the number of satisfied constraints.\\
$\bullet$~\textbf{Avg Tool Calls} (ATC): the average number of tool invocations per instance during inference.


\subsection{Main Results}
As shown in Table~\ref{tab:main_results}, COSMO-Agent (8B) achieves an FSR of 74.5\%, which is the best among all methods. Compared with the strongest open-source baseline intern-S1 (32.0\%), it improves by 42.5\%. Compared with the best closed-source baseline Gemini-3-Flash (67.5\%), it improves by 7.0\%. This indicates that COSMO-Agent can more reliably produce feasible solutions that simultaneously satisfy the displacement, stress, and cost constraints. In terms of constraint satisfaction, COSMO-Agent achieves a displacement satisfaction rate of 87.5\%, a stress satisfaction rate of 76.0\%, and a cost satisfaction rate of 93.5\%, which are overall the best. For structured outputs, COSMO-Agent reaches an MEO of 100\%, ensuring that the final results can be parsed into an executable JSON and thus improving the reliability of end-to-end reproduction. From the optimization process perspective, COSMO-Agent obtains an AS of 0.6504, slightly lower than Gemini's 0.6802. Since AS includes tool-call rewards, Gemini uses more tool calls per successful case on average (9.32), making it easier to accumulate process scores. In contrast, COSMO-Agent has an ATC of 6.72, indicating that it requires fewer tool calls to reach feasible solutions and is more interaction-efficient. Overall, COSMO-Agent achieves high success rates while maintaining good inference efficiency.

\begin{table}[t]
\centering
\caption{Main results on the test set. Abbreviations: \textbf{FSR} (Full Success Rate), \textbf{DSR} (Displacement Satisfaction Rate), \textbf{SSR} (Stress Satisfaction Rate), \textbf{CSR} (Cost Satisfaction Rate), \textbf{MEO} (Model Extract Output), \textbf{AS} (Average Score), \textbf{ATC} (Avg Tool Calls).}
\label{tab:main_results}
\begin{adjustbox}{width=\columnwidth,center}
\setlength{\tabcolsep}{3.8pt}
\renewcommand{\arraystretch}{1.05}
\begin{tabular}{l r r r r r r r r}
\hline
\textbf{Model} & \textbf{Scale} & \textbf{FSR} & \textbf{DSR} & \textbf{SSR} & \textbf{CSR} & \textbf{MEO} & \textbf{AS} & \textbf{ATC} \\
\hline
Intern-S1-mini & 8B   & 20.0\% & 24.0\% & 31.5\% & 32.5\% & 40.0\%  & 0.2820 & 6.31 \\
Llama-4-Scout  & 17B  & 21.0\% & 31.5\% & 42.0\% & 45.5\% & 62.5\%  & 0.2689 & \textbf{2.94} \\
Qwen3-30B      & 30B  & 29.5\% & 48.5\% & 74.5\% & 73.0\% & \textbf{100.0\%} & 0.5789 & 8.60 \\
Qwen3-Next     & 80B  & 25.5\% & 47.5\% & 75.0\% & 58.5\% & 99.5\%  & 0.5630 & 8.60 \\
Intern-S1      & 236B & 32.0\% & 53.0\% & 75.0\% & 60.0\% & 99.5\%  & 0.5367 & 7.44 \\
\hline
Claude-Sonnet-4.5 & -- & 36.0\% & 56.0\% & 70.5\% & 74.5\% & 92.5\% & 0.4809 & 11.25 \\
Gemini-3-Flash    & -- & 67.5\% & 83.0\% & 75.0\% & 91.0\% & 98.0\% & \textbf{0.6802} & 9.32 \\
\hline
\textbf{COSMO-Agent} & 8B & \textbf{74.5\%} & \textbf{87.5\%} & \textbf{76.0\%} & \textbf{93.5\%} & \textbf{100.0\%} & 0.6504 & 6.72 \\
\hline
\end{tabular}
\end{adjustbox}
\end{table}

\begin{table}[t]
\centering
\caption{Generalization results on the unseen-category set.}
\label{tab:generalization_results}
\begin{adjustbox}{width=\columnwidth,center}
\setlength{\tabcolsep}{3.8pt}
\renewcommand{\arraystretch}{1.05}
\begin{tabular}{l r r r r r r r r}
\hline
\textbf{Model} & \textbf{Scale} & \textbf{FSR} & \textbf{DSR} & \textbf{SSR} & \textbf{CSR} & \textbf{FE} & \textbf{AS} & \textbf{ATC} \\
\hline
Intern-S1-mini & 8B   & 20.0\% & 27.0\% & 41.0\% & 31.0\% & 49.0\%  & 0.3111 & 5.81 \\
Llama-4-Scout  & 17B  & 19.0\% & 33.0\% & 43.0\% & 37.0\% & 62.0\%  & 0.2520 & \textbf{3.44} \\
Qwen3-30B      & 30B  & 28.0\% & 45.0\% & \textbf{81.0\%} & 71.0\% & \textbf{100.0\%} & 0.5870 & 8.41 \\
Qwen3-Next     & 80B  & 24.0\% & 43.0\% & 80.0\% & 57.0\% & \textbf{100.0\%} & 0.5690 & 8.71 \\
Intern-S1      & 236B & 35.0\% & 53.0\% & 79.0\% & 63.0\% & 99.0\%  & 0.5688 & 7.56 \\
\hline
Claude-Sonnet-4.5 & -- & 38.0\% & 53.0\% & 81.0\% & 73.0\% & 94.0\% & 0.6468 & 11.08 \\
Gemini-3-Flash    & -- & 57.0\% & 60.0\% & 57.0\% & 60.0\% & 60.0\% & \textbf{0.6977} & 9.44 \\
\hline
\textbf{COSMO-Agent} & 8B & \textbf{75.0\%} & \textbf{84.0\%} & 78.0\% & \textbf{89.0\%} & \textbf{100.0\%} & 0.6150 & 6.57 \\
\hline
\end{tabular}
\end{adjustbox}
\end{table}

\subsection{Generalization Performance}
As shown in Table~\ref{tab:generalization_results}, on the generalization set consisting of five unseen categories, COSMO-Agent achieves an FSR of 75.0\%, which is broadly consistent with the main-test result (74.5\%), indicating no obvious degradation on unseen templates. Compared with the baselines, COSMO-Agent remains significantly ahead: among open-source methods, intern-S1 achieves 35.0\%; among closed-source methods, Gemini-3-Flash achieves 57.0\% and Claude-Sonnet-4.5 achieves 38.0\%.

On the generalization set, COSMO-Agent still maintains a 100\% format extraction success rate, ensuring stable end-to-end reproducibility. In contrast, Gemini-3-Flash has a format extraction success rate of only 60.0\%, which directly reduces the effective end-to-end success proportion in evaluation. From the optimization process perspective, COSMO-Agent obtains an AS of 0.6150, which is lower than Gemini (0.6977) and Claude (0.6468). This difference is consistent with the AS scoring rule: AS accumulates tool-call rewards, and Gemini and Claude use more tool calls per successful case on average (9.44 and 11.08, respectively) than COSMO-Agent (6.57), making it easier to obtain higher process scores. Nevertheless, on the more critical end-to-end metric FSR, COSMO-Agent still maintains a clear advantage and also demonstrates high interaction efficiency on unseen categories.

\begin{figure}[t]
  \centering
  \includegraphics[width=\columnwidth]{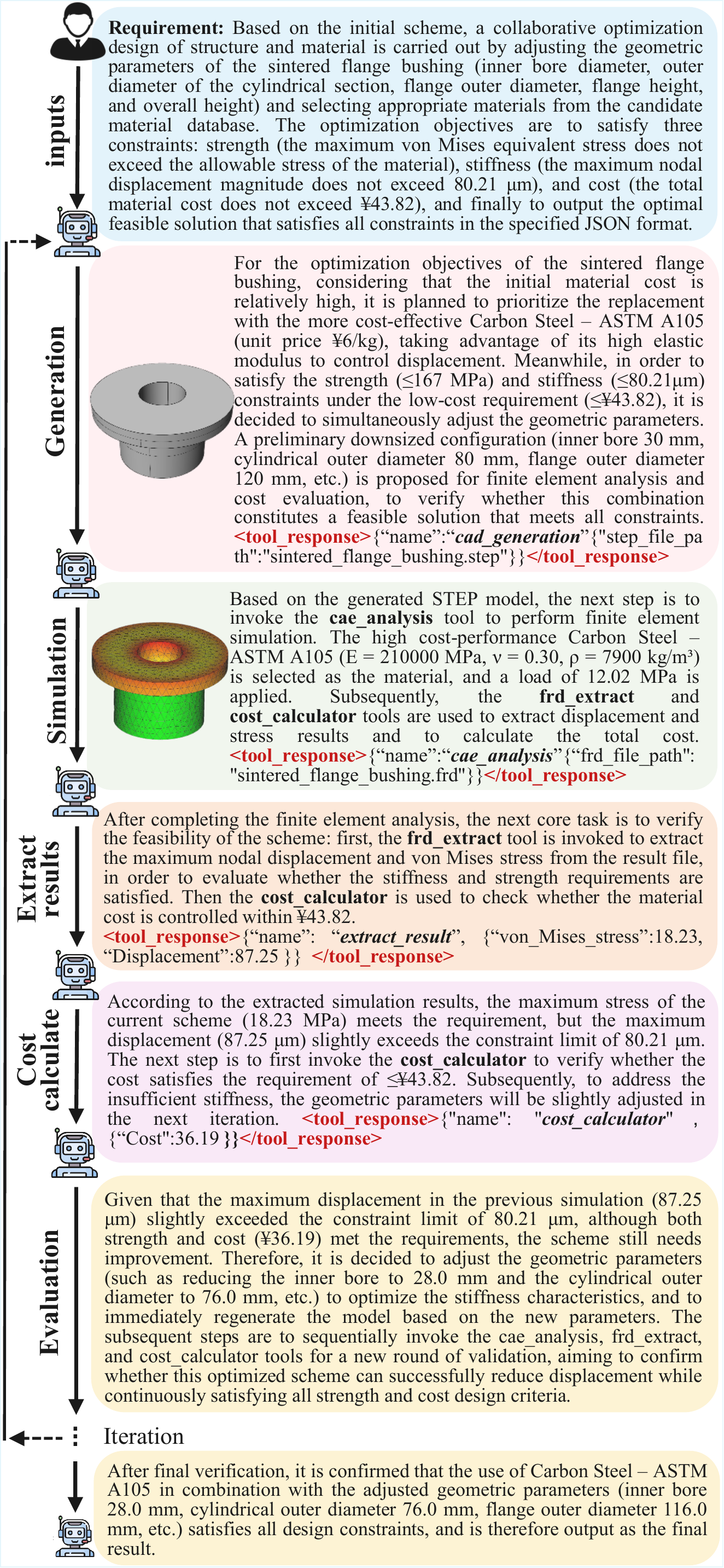}

  \caption{Visualized inference cases of COSMO-Agent.}
  \label{fig:case_visual}
\end{figure}

\subsection{Visualization Result}
Fig.\ref{fig:case_visual} illustrates \textsc{COSMO-Agent}'s dynamic reasoning through a structure--material co-optimization of a sintered flange bushing. To meet the cost constraint, it first switches the material to Carbon Steel--ASTM A105 and proposes a reduced-size geometry, then iteratively invokes CAD generation, CAE simulation, result extraction, and cost calculation in a closed-loop ``generation--simulation--evaluation'' workflow. The first evaluation shows stress meeting the strength constraint, but displacement (87.25~$\mu$m) slightly exceeding the stiffness limit (80.21~$\mu$m), while cost (\textyen36.19) remains within budget. \textsc{COSMO-Agent} therefore improves stiffness via minor geometry updates and re-verifies, and the final design satisfies strength, stiffness, and cost constraints.



\subsection{Ablation Studies}
We conduct ablation studies to identify the key factors behind the performance gains of COSMO-Agent: (i) whether GRPO-based RL training is applied, and (ii) whether the reward is computed from toolchain rollout logs. All other settings are kept identical, and the results are summarized in Table~\ref{tab:ablation_results}.

\paragraph{Effect of RL training.}
Without RL (w/o RL), the model achieves an FSR of 26.0\%. With RL training and the full reward, the FSR increases to 74.5\%, improving by 48.5 percentage points. The displacement satisfaction rate improves from 39.5\% to 87.5\%, the cost satisfaction rate improves from 65.0\% to 93.5\%, and the stress satisfaction rate improves from 72.0\% to 76.0\%. These results indicate that RL training substantially strengthens the model's ability to update parameters based on numerical feedback in closed-loop interactions.

\paragraph{Effect of rollout-log-based reward.}
We compare our rollout-log-based reward with a ``final-JSON re-verification'' baseline, where the reward is computed by parsing the model's final JSON output, re-running the CAD--CAE toolchain once with the reported $(\mathbf{p}, m)$, and then scoring feasibility based on the re-simulated metrics (rather than using the rollout interaction logs).
Replacing our reward with this baseline yields an FSR of 36.0\%, which is much lower than the full COSMO-Agent (74.5\%). Under this setting, the average tool calls drop to 2.62. By inspecting rollout logs, we find that the model tends to avoid calling tools and directly outputs a guessed JSON solution, weakening closed-loop optimization. In contrast, our rollout-log-based reward directly parses tool-interaction logs and uses the last complete metric triple to compute returns, avoiding the high cost of re-simulation and more effectively encouraging the model to follow the ``call tools--read feedback--iterate'' loop.

\begin{table}[t]
\centering
\caption{Ablation studies on the test set. ``w/o RL'' disables GRPO training. ``w/o Rollout Reward'' replaces the rollout-log-based reward with a final-JSON re-verification reward.}
\label{tab:ablation_results}
\begin{adjustbox}{width=\columnwidth,center}
\begin{tabular}{l r r r r r r r}
\hline
\textbf{Setting} & \textbf{FSR} & \textbf{DSR} & \textbf{SSR} & \textbf{CSR} & \textbf{MEO} & \textbf{AS} & \textbf{ATC} \\
\hline
w/o RL & 26.0\% & 39.5\% & 72.0\% & 65.0\% & 98.5\% & 0.4906 & 6.08 \\
w/o Rollout Reward & 36.0\% & 59.0\% & 54.0\% & 69.0\% & 100.0\% & 0.3760 & \textbf{2.62} \\
\textbf{COSMO-Agent} & \textbf{74.5\%} & \textbf{87.5\%} & \textbf{76.0\%} & \textbf{93.5\%} & \textbf{100.0\%} & \textbf{0.6504} & 6.72 \\
\hline
\end{tabular}
\end{adjustbox}
\end{table}

%% file: sec/conclusion.tex
\section{Conclusion}

We presented COSMO-Agent, a tool-augmented reinforcement learning framework for reliable closed-loop CAD--CAE iteration. By modeling the CAD--CAE pipeline as an interactive environment and training an 8B LLM with GRPO and a rollout-log-based reward, COSMO-Agent learns to generate structured, executable parametric edits and improve designs through multi-round tool feedback. The rollout-log reward leverages tool execution logs to encourage proper tool use and constraint-driven optimization without requiring additional expensive re-simulations.
Experiments show that COSMO-Agent significantly improves feasibility and interaction efficiency under fixed tool-call and retry budgets, and it generalizes well to unseen component categories. 

In future work, we will extend COSMO-Agent to richer design settings with contact, assembly, and multi-part constraints, and to additional physics such as nonlinear materials and coupled multi-physics. We also plan to support alternative CAD and CAE backends and study scalability under larger action spaces, tighter budgets, and more diverse failure modes. Finally, we will explore improved training curricula and robustness objectives to further strengthen long-horizon reliability in practical pipelines.